\def\year{2020}\relax
\documentclass[letterpaper]{article} 
\usepackage{aaai20}  
\usepackage{times}  
\usepackage{helvet} 
\usepackage{courier}  
\usepackage[hyphens]{url}  
\usepackage{graphicx} 
\usepackage{graphicx}  
\usepackage{subfigure}
\usepackage{booktabs}
\usepackage[namelimits]{amsmath} 
\usepackage{amssymb}             
\usepackage{amsfonts}            
\usepackage{mathrsfs}            
\usepackage{verbatim}

\title{Improving Adversarial Robustness via Attention and Adversarial Logit Pairing}
\author{Dou Goodman$^{\dag}$${}^{*}$, Xingjian Li${\ddag}$${}^{*}$, Ji Liu${\ddag}$, Dejing Dou${\ddag}$, Tao Wei$^{\dag}$\\
$^\dag$ X-Lab, Baidu Inc.\\
$\ddag$ Big Data Lab, Baidu Reaseach\\
\texttt{\{liu.yan,lixingjian,liuji04,doudejing,lenx\}@baidu.com} \\
\thanks{\textbf{${}^{*}$Authors colloborated equally and are ordered by their last names.}}}

\begin{document}

\maketitle

\begin{abstract}
Though deep neural networks have achieved the state of the art performance in visual classification, recent studies have shown that they are all vulnerable to the attack of adversarial examples. In this paper, we develop improved techniques for defending against adversarial examples. First, we propose an enhanced defense technique denoted \textbf{Attention and Adversarial Logit Pairing(AT+ALP)}, which encourages both attention map and logit for the pairs of examples to be similar. When being applied to clean examples and their adversarial counterparts, \textbf{AT+ALP} improves accuracy on adversarial examples over adversarial training. We show that \textbf{AT+ALP} can effectively increase the average activations of adversarial examples in the key area and demonstrate that it focuses on discriminate features to improve the robustness of the model. Finally, we conduct extensive experiments using a wide range of datasets and the experiment results show that our \textbf{AT+ALP} achieves \textbf{the state of the art} defense performance. For example, on \textbf{17 Flower Category Database}, under strong 200-iteration \textbf{PGD} gray-box and black-box attacks where prior art has 34\% and 39\% accuracy, our method achieves \textbf{50\%} and \textbf{51\%}. Compared with previous work, our work is evaluated under highly challenging PGD attack: the maximum perturbation $\epsilon \in \{0.25,0.5\}$ i.e. $L_\infty \in \{0.25,0.5\}$ with 10 to 200 attack iterations. To the best of our knowledge, such a strong attack has not been previously explored on a wide range of datasets.

\end{abstract}

\section{Introduction}
In recent years, deep neural networks have been extensively deployed for computer vision tasks, particularly for visual classification problems, where new algorithms have been reported to achieve even better performance than human beings \cite{Krizhevsky2012ImageNet,DBLP:journals/corr/HeZRS15,li2019dsfd}. The success of deep neural networks has led to an explosion in demand. However, recent studies have shown that they are all vulnerable to the attack of adversarial examples \cite{szegedy2013intriguing,Carlini2016Towards,moosavi2016deepfool,Bose2018Adversarial}.
Small and often imperceptible perturbations to the input images are sufficient to fool the most powerful deep neural networks.

\begin{figure}[t]
	\centering
	
	\subfigure[]{%
			\includegraphics[width=0.8in]{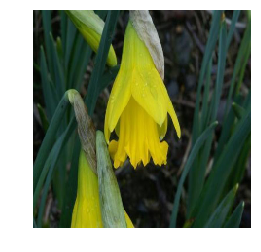}	
			\label{fig:o}
	}%
	\subfigure[]{%
			\includegraphics[width=0.8in]{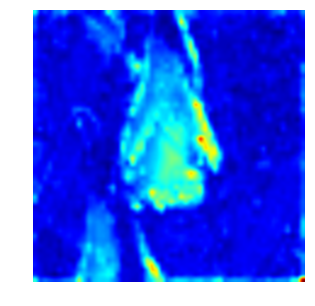}	
			\label{fig:o-group01}
	}%
	\subfigure[]{%
			\includegraphics[width=0.8in]{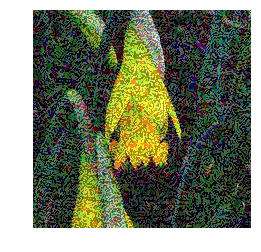}	
			\label{fig:adv}
	}%
	\subfigure[]{%
			\includegraphics[width=0.8in]{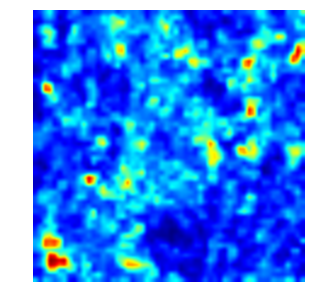}	
			\label{fig:adv-group01}
	}%
	
	\centering
	\caption{(a) is original image and (b) is corresponding spatial attention map of ResNet-50\cite{DBLP:journals/corr/HeZRS15} pretrained on ImageNet\cite{ILSVRC15} which shows where the network focuses in order to classify the given image.(c) is adversarial image of (a) , (d) is corresponding spatial attention map. }
	\label{fig:introduce}
\end{figure}

%
In Figure \ref{fig:introduce}, we visualize the spatial attention map of a flower and its corresponding adversarial image on ResNet-50 \cite{DBLP:journals/corr/HeZRS15} pretrained on ImageNet\cite{ILSVRC15}. The figure suggests that adversarial perturbations, while small in the pixel space, lead to very substantial “noise” in the attention map of the network. Whereas the features for the clean image appear to focus primarily on semantically informative content in the image, the attention map for the adversarial image are activated across semantically irrelevant regions as well.
The state of the art adversarial training methods only encourage hard labels \cite{Madry2017Towards,tramer2017ensemble} or logit \cite{DBLP:journals/corr/abs-1803-06373} for pairs of clean examples and adversarial counterparts to be similar. In our opinion, it is not enough to align the difference between the clean examples and adversarial counterparts only at the end part of the whole network,i.e., hard labels or logit, and we need to align the attention maps for important parts of the whole network.
Motivated by this observation, we explore \textbf{Attention and Adversarial Logit Pairing(AT+ALP)}, a method that encourages both attention map and logit for pairs of examples to be similar. When being applied to clean examples and their adversarial counterparts, \textbf{AT+ALP} improves accuracy on adversarial examples over adversarial training.

The contributions of this paper are summarized as follows:

\begin{itemize}
\item We introduce enhanced adversarial training using a technique we call \textbf{Attention and Adversarial Logit Pairing(AT+ALP)}, which encourages both attention map and logit for pairs of examples to be similar. When being applied to clean examples and their adversarial counterparts, \textbf{AT+ALP} improves accuracy on adversarial examples over adversarial training.
\item We show that our \textbf{AT+ALP} can effectively increase the average activations of adversarial examples in the key area and demonstrate that it focuses on more discriminate features to improve the robustness of the model.
\item We show that our \textbf{AT+ALP} achieves \textbf{the state of the art} defense on a wide range of datasets against strong \textbf{PGD} gray-box and black-box attacks. Compared with previous work,our work is evaluated under highly challenging PGD attack: the maximum perturbation $\epsilon \in \{0.25,0.5\}$, i.e., $L_\infty \in \{0.25,0.5\}$ with 10 to 200 attack iterations. To the best of our knowledge, such a strong attack has not been previously explored on a wide range of datasets.
\end{itemize}

The rest of the paper is organized as follows: in Section \ref{sec:RelatedWork}2, we present the related works; in Section \ref{sec:Methods}3, we introduce definitions and threat models; in Section \ref{sec:Methods}4 we propose our \textbf{Attention and Adversarial Logit Pairing(AT+ALP)} method; in Section \ref{sec:Experiments}5, we show extensive experimental results; and Section \ref{sec:Conclusion}6 concludes.

\section{Related Work}

\label{sec:RelatedWork}

\cite{obfuscated-gradients} evaluate the robustness of nine papers \cite{buckman2018thermometer,ma2018characterizing,guo2017countering,dhillon2018stochastic,Xie2017Mitigating,song2017pixeldefend,samangouei2018defense,Madry2017Towards,na2017cascade} accepted by ICLR 2018 as non-certified white-box-secure defenses to adversarial examples. They find that seven of the nine defenses use obfuscated gradients, a kind of gradient masking, as a phenomenon that leads to a false sense of security in defenses against adversarial examples. Obfuscated gradients provide a limited increase in robustness and can be broken by improved attack techniques they develop. The only defense they observe that significantly increases robustness to adversarial examples within the threat model proposed is \textbf{adversarial training}\cite{Madry2017Towards}.

Adversarial training \cite{Goodfellow2014Explaining,Madry2017Towards,DBLP:journals/corr/abs-1803-06373,tramer2017ensemble,Pang2019Improving} defends against adversarial perturbations by training networks on adversarial images that are generated on-the-fly during training. For adversarial training, the most relevant work to our study is \cite{DBLP:journals/corr/abs-1803-06373}, which introduce a technique they call \textbf{Adversarial Logit Pairing(ALP)}. This method encourages logits for pairs of examples to be similar. Our \textbf{AT+ALP} encourages both attention map and logit for pairs of examples to be similar. When being applied to clean examples and their adversarial counterparts, \textbf{AT+ALP} improves accuracy on adversarial examples over adversarial training. \cite{Araujo2019Robust} adds random noise at training and inference time, \cite{Xie2018Feature} adds denoising blocks to the model to increase adversarial robustness, while neither of the above approaches focuses on the attention map.

In terms of methodologies, our work is also related to deep transfer learning and knowledge distillation problems, and the most relevant work to our study is \cite{Zagoruyko2016Paying,li2019delta}, which constrain the $L_2$-norm of the difference between their behaviors (i.e., the feature maps of outer layer outputs in the source/target networks). Our \textbf{AT+ALP} constrains attention map and logit for pairs of clean examples and their adversarial counterparts to be similar.

\section{Definitions and Threat Models}
\label{sec:DefinitionsThreatModels}

In this paper, we always assume the attacker is capable of forming attacks that consist of perturbations of limited $L_\infty$-norm. This is a simplified task chosen because it is more amenable to benchmark evaluations. We consider two different threat models characterizing amounts of information the adversary can have:

\begin{itemize}
\item \textbf{Gray-box Attack} We focus on defense against gray-box attacks in this paper. In a gray-back attack, the attacker knows both the original network and the defense algorithm. Only the parameters of the defense model are hidden from the attacker. This is also a standard setting assumed in many security systems and applications \cite{Pfleeger2004Security}.
\item \textbf{Black-box Attack} The attacker has no information about the model’s architecture or parameters, and no ability to send queries to the model to gather more information.
\end{itemize}

\section{Methods}
\label{sec:Methods}

\subsection{Architecture}

\begin{figure*}[htbp]
	\centering
	\includegraphics[width=5in]{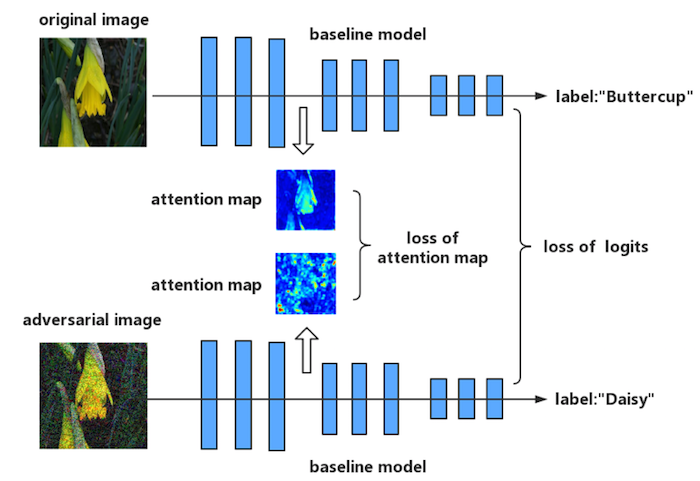}	
	\centering
	\caption{Schematic representation of \textbf{Attention and Adversarial Logit Pairing(AT+ALP)}: a baseline model is trained so as, not only to make similar logits, but to also have similar spatial attention maps to those of original image and adversarial image. }
	\label{fig:architectural}
\end{figure*}

Figure \ref{fig:architectural} represents architecture of \textbf{Attention and Adversarial Logit Pairing(AT+ALP)}: a baseline model is adversarial trained so as, not only to make similar logits, but to also have similar spatial attention maps to those of original image and adversarial image. 

\subsection{Adversarial training}
We use adversarial training with \textbf{Projected Gradient Descent(PGD)}\cite{Madry2017Towards} as the underlying basis for our methods:
\begin{equation}
\underset{\theta}{\arg \min } \mathbb{E}_{(x, y) \in \hat{p}_{\text { data }}}\left(\max _{\delta \in S} L(\theta, x+\delta, y)\right)
\end{equation}
where $\hat{p}_{\text { data }}$ is the underlying training data distribution, $L(\theta, x+\delta, y)$ is a loss function at data point $x$ which has true class $y$ for a model with parameters $\theta$, and the maximization with respect to $\delta$ is approximated using PGD.
In this paper,the loss is defined as:
\begin{equation}
L=L_{CE}+\alpha L_{ALP}+\beta L_{AT},
\end{equation}
where $L_{CE}$ is cross entropy, $\alpha$ and $\beta$ are hyperparameters.

\subsection{Adversarial Logit Pairing}
We also use \textbf{Adversarial Logit Pairing(ALP)} to encourage the logits from clean examples and their adversarial counterparts to be similar to each other. For a model that takes inputs $x$ and computes a vector of logit $z = f(x)$, logit pairing adds a loss:
\begin{equation}
L_{ALP}=L_a(f(x),f(x+\delta))
\end{equation}

In this paper we use $L_2$ loss for $L_a$.

\subsection{Attention Map}
We  use \textbf{Attention Map(AT)} to encourage the attention map from clean examples and their adversarial counterparts to be similar to each other. Let $I$ denote the indices of all activation layer pairs, for which we want to pay attention. Then, we can define the following total loss:
\begin{equation}
L_{AT}=\sum_{j \in \mathcal{I}}\left\|\frac{Q_{ADV}^{j}}{\left\|Q_{ADV}^{j}\right\|_{2}}-\frac{Q_{O}^{j}}{\left\|Q_{O}^{j}\right\|_{2}}\right\|_{p} 
\end{equation}
Let $O$, $ADV$ denote clean examples and their adversarial counterparts, where $Q_{O}^{j}=\operatorname{vec}\left(F\left(A_{O}^{j}\right)\right)$ and $Q_{ADV}^{j}=\operatorname{vec}\left(F\left(A_{ADV}^{j}\right)\right)$ are respectively the j-th pair of clean examples and
their adversarial counterparts attention maps in vectorized form, and $p$ refers to norm type (in the experiments we use $p = 2$). 

\section{Experiments:Gray and Black-Box Settings}
\label{sec:Experiments}
To evaluate the effectiveness of our defense strategy, we performed a series of image-classification experiments on \textbf{17 Flower Category Database}\cite{Nilsback06}, \textbf{Part of ImageNet Database} and \textbf{Dogs-vs-Cats Database}. Following \cite{obfuscated-gradients,Xie2018Feature}, we assume an adversary that uses the state of the art PGD adversarial attack method.

We consider untargeted attacks when evaluating under the gray and black-box settings; untargeted attacks are also used in our adversarial training. We evaluate top-1 classification accuracy on validation images that are adversarially perturbed by the attacker. In this paper, adversarial perturbation is considered under $L_\infty$ norm (i.e., maximum perturbation for each pixel), with an allowed maximum value of $\epsilon$. The value of $\epsilon$ is relative to the pixel intensity scale of 256,we use $\epsilon=64/256=0.25$ and $\epsilon=128/256=0.5$. PGD attacker with 10 to 200 attack iterations and step size $\alpha=1.0/256=0.0039$.
Our baselines are ResNet-101/152. There are four groups of convolutional structures in the baseline
model, group-0 extracts of low-level features,group-1 and group-2 extract of mid-level features,group-3 extracts of high-level features\cite{DBLP:journals/corr/ZagoruykoK16a}, which are described as $conv2\_x$, $conv3\_x$,$conv4\_x$ and $conv5\_x$  in \cite{DBLP:journals/corr/HeZRS15}

\begin{figure*}[htb]
	\centering
	
	\subfigure[]{%
			\includegraphics[width=3in]{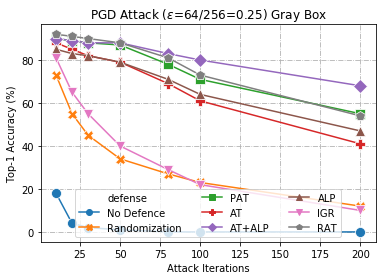}	
			\label{fig:flowers17:graybox-64}
	}%
	\subfigure[]{%
			\includegraphics[width=3in]{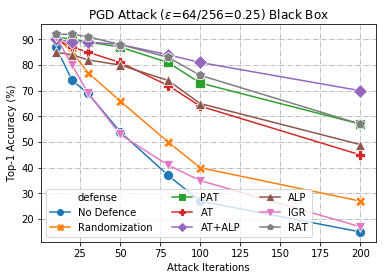}	
			\label{fig:flowers17:blackbox-64}
	}%
	
	\subfigure[]{%
			\includegraphics[width=3in]{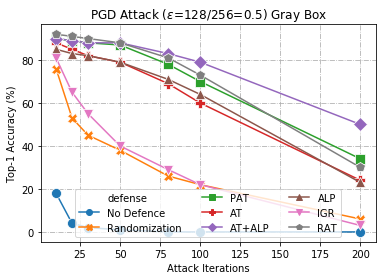}	
			\label{fig:flowers17:graybox-128}
	}%
	\subfigure[]{%
			\includegraphics[width=3in]{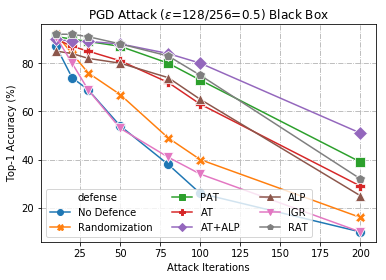}	
			\label{fig:flowers17:blackbox-128}
	}%
	
	\centering
	\caption{\textbf{Defense against gray-box and black-box attacks on 17 Flower Category Database}.(a)(c) shows results against a gray-box PGD attacker with 10 to 200 attack iterations.(b)(d) shows results against a black-box PGD attacker with 10 to 200 attack iterations.The maximum perturbation is $\epsilon \in \{0.25,0.5\}$ i.e. $L_\infty \in \{0.25,0.5\}$.Our \textbf{AT+ALP}(purple line) outperform the state-of-the-art in adversarial robustness against highly challenging gray-box and black-box PGD attacks.}
	\label{fig:flowers17:defend-64}
\end{figure*}

\subsection{Image Database}

We performed a series of image-classification experiments on a wide range of datasets.
\begin{itemize}
\item \textbf{17 Flower Category Database}\cite{Nilsback06} contains images of flowers belonging to 17 different categories.The images were acquired by searching the web and taking pictures. There are 80 images for each category.
\item \textbf{Part of ImageNet Database} contains images of four objects. These four objects are randomly selected from the ImageNet Database\cite{ILSVRC15}. In this experiment, they are tench, goldfish, white shark and dog. Each object contains 1300 training images and 50 test images.
\item \textbf{Dogs-vs-Cats Database}\footnote{https://www.kaggle.com/chetankv/dogs-cats-images} contains 8,000 images of dogs and cats in the train dataset and 2,000 in the test val dataset.
\end{itemize}

\subsection{Experimental Setup}

To perform image classification, we use ResNet-101/152 that were trained on the \textbf{17 Flower Category Database},\textbf{Part of ImageNet Database} and \textbf{Dogs-vs-Cats Database} training set. We consider two different attack settings: (1) a gray-box attack setting in which the model used to generate the adversarial images is the same as the image- classification model, viz. the ResNet-101; and (2) a black-box attack setting in which the adversarial images are generated using the ResNet-152 model;The backend prediction model of gray-box and black-box is ResNet-101 with different implementations of the state of the art defense methods,such as IGR\cite{Ross2017Improving},PAT\cite{Madry2017Towards},RAT\cite{Araujo2019Robust},Randomization\cite{Xie2017Mitigating},ALP\cite{DBLP:journals/corr/abs-1803-06373},FD\cite{Xie2018Feature} and ADP\cite{Pang2019Improving}.

\begin{figure*}[!htb]
	\centering
	\subfigure[original]{%
			\includegraphics[width=1in]{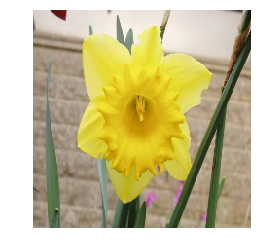}	
			\label{fig:flowers17:o}
	}%
	\subfigure[adversarial]{%
			\includegraphics[width=1in]{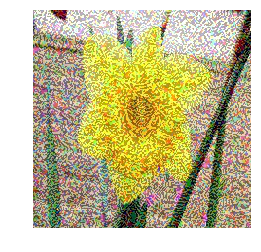}	
			\label{fig:flowers17:adv}
	}%
	
	\subfigure[Baseline model,labeled as "Buttercup"]{%
			\includegraphics[width=3.5in]{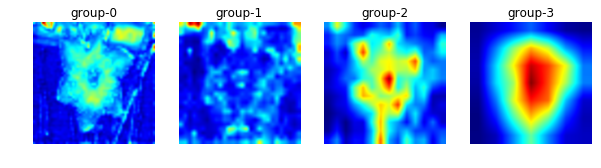}	
			\label{fig:flowers17:o-std}
	}%
	\subfigure[Baseline model,labeled as "Windflower"]{%
			\includegraphics[width=3.5in]{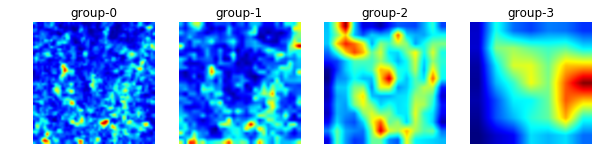}	
			\label{fig:flowers17:adv-std}
	}%
	
	\subfigure[ALP,labeled as "Bluebell"]{%
			\includegraphics[width=3.5in]{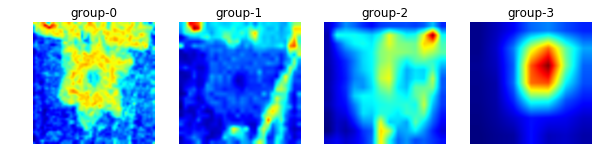}	
			\label{fig:flowers17:adv-alp}
	}%
	\subfigure[AT,labeled as "LilyValley"]{%
			\includegraphics[width=3.5in]{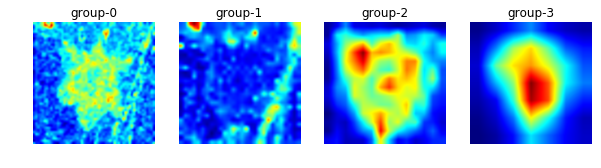}	
			\label{fig:flowers17:adv-at}
	}%
	
	\subfigure[AT+ALP,labeled as "Buttercup"]{%
			\includegraphics[width=3.5in]{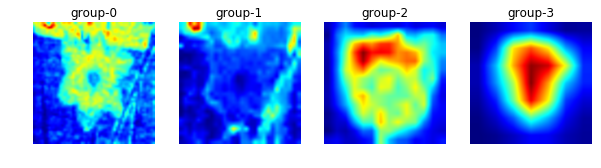}	
			\label{fig:flowers17:adv-at-alp}
	}%

	\centering
	\caption{\textbf{Activation attention maps for defense against gray-box PGD attacks($\epsilon= 0.25$) on 17 Flower Category Database}.(a) is original image and (b) is corresponding adversarial image.(c) and (d) are activation attention maps of baseline model for original image and adversarial image, (e)-(f) are activation attention maps of \textbf{ALP},\textbf{AT} and \textbf{AT+ALP} for adversarial image.Group-0 to group-3 represent the activation attention maps of four groups of convolutional structures in the baseline model,group-0 extracts of low-level features, group-1 and group-2 extract of mid-level features, group-3 extracts of high-level features\cite{DBLP:journals/corr/ZagoruykoK16a}.
	\textbf{It can be clearly found that group-0 of AT + ALP can extract the outline and texture of flowers more accurately, and group-3 has a higher level of activation on the whole flower, compared with other defense methods, only it makes accurate prediction.}}
	\label{fig:flowers17:attention}
\end{figure*}

\begin{table*}[htbp]
\caption{\textbf{Defense against gray-box and black-box attacks on 17 Flower Category Database, Part of ImageNet Database and Dogs-vs-Cats Database}. The adversarial perturbation were produced using PGD with step size $\alpha=1.0/256=0.0039$ and 200 attack iterations.As shown in this table, \textbf{AT+ALP got the highest Top-1 Accuracy on all these database}. }
\label{tab:main_result}
~\\
\centering
\resizebox{0.6\textwidth}{!}{
\begin{tabular}{l|cc|cc}
\textbf{17 Flower Category Database}& \multicolumn{2}{l}{Gray-Box} & \multicolumn{2}{l}{Black-Box} \\
$\epsilon=L_\infty$&       0.25  & 0.5 &       0.25 & 0.5 \\
\midrule
No Defence    &         0 &   0 &        15 &  10 \\
\midrule
IGR\cite{Ross2017Improving}           &        10 &   3 &        17 &  10 \\
PAT\cite{Madry2017Towards}           &        55 &  34 &        57 &  39 \\
RAT\cite{Araujo2019Robust}           &        54 &  30 &        57 &  32 \\
Randomization\cite{Xie2017Mitigating} &        12 &   6 &        27 &  16 \\
ALP\cite{DBLP:journals/corr/abs-1803-06373}           &        47 &  23 &        49 &  25 \\
FD\cite{Xie2018Feature}           &        33&  10 &        33 &  10 \\
ADP\cite{Pang2019Improving}           &        22&  8 &        23 &  8 \\
Our \textbf{AT}            &        41 &  24 &        45 &  29 \\
\midrule
Our \textbf{AT+ALP}         &        \textbf{68} &  \textbf{50} &        \textbf{70} &  \textbf{51} \\
\end{tabular}
}

~\\
~\\

\resizebox{0.6\textwidth}{!}{
\begin{tabular}{l|cc|cc}
\textbf{Part of ImageNet Database}& \multicolumn{2}{l}{Gray-Box} & \multicolumn{2}{l}{Black-Box} \\
$\epsilon=L_\infty$&       0.25  & 0.5 &       0.25 & 0.5 \\
\midrule
No Defence    &         2 &   3 &        52 &  50 \\
\midrule
IGR\cite{Ross2017Improving}           &        32 &  32 &        34 &  34 \\
PAT\cite{Madry2017Towards}           &        76 &  76 &        77 &  77 \\
RAT\cite{Araujo2019Robust}           &        76 &  76 &        77 &  76 \\
Randomization\cite{Xie2017Mitigating} &        40 &  41 &        62 &  59 \\
ALP\cite{DBLP:journals/corr/abs-1803-06373}           &        54 &  54 &        55 &  55 \\
FD\cite{Xie2018Feature}           &        60&  61 &        61 &  61\\
ADP\cite{Pang2019Improving}           &        42&  44 &        43 &  44 \\
Our \textbf{AT}            &        76 &  76 &        77 &  76 \\
\midrule
Our \textbf{AT+ALP}         &        \textbf{82} &  \textbf{82} &        \textbf{82} &  \textbf{82} \\
\end{tabular}
}

~\\
~\\

\resizebox{0.6\textwidth}{!}{
\begin{tabular}{l|cc|cc}
\textbf{Dogs-vs-Cats Database}& \multicolumn{2}{l}{Gray-Box} & \multicolumn{2}{l}{Black-Box} \\
$\epsilon=L_\infty$&       0.25  & 0.5 &       0.25 & 0.5 \\
\midrule
No Defence    &         1 &   1 &        52 &  53\\
\midrule
IGR\cite{Ross2017Improving}           &        57 &  60 &        51 &  52 \\
PAT\cite{Madry2017Towards}           &        51 &  51 &        52 &  52 \\
RAT\cite{Araujo2019Robust}           &        49 &  49 &        50 &  50 \\
Randomization\cite{Xie2017Mitigating} &        10 &   8 &        55 &  54 \\
ALP\cite{DBLP:journals/corr/abs-1803-06373}           &        57 &  56 &        57 &  57 \\
FD\cite{Xie2018Feature}           &        57&  57 &        57 &  57\\
ADP\cite{Pang2019Improving}           &        50&  50 &        50 &  50 \\
Our \textbf{AT}            &        50 &  50 &        50 &  50 \\
\midrule
Our \textbf{AT+ALP}         &        \textbf{67} &  \textbf{67} &        \textbf{71} &  \textbf{71} \\
\end{tabular}
}

\end{table*}

\begin{table*}[htb]
\caption{Comparing \textbf{average activations} on discriminate parts of \textbf{17 Flower Category Database} for different defense methods.In addition, we included new statistical results of activations on part locations of \textbf{17 Flower Category Database} supporting the above qualitative cases. The \textbf{17 Flower Category Database} defined  discriminative parts of flowers. So for each image, we got several key  regions which are very important to discriminate its category. Using all testing examples of \textbf{17 Flower Category Database}, we calculated normalized activations on these key regions of these different defense methods. As shown in this table, \textbf{AT+ALP} got the highest average activations on those key regions, demonstrating that \textbf{AT+ALP} focused on more discriminate features for flowers recognition.}
\label{tab:flower17:at_loss}
~\\
\centering
\resizebox{0.65\textwidth}{!}{
\begin{tabular}{l|cc|cc}

Defense & \multicolumn{2}{c}{Black-Box} & \multicolumn{2}{c}{Gray-Box} \\
$\epsilon=L_\infty$ &                 $0.25$  &       $0.5$ &       $0.25$  &       $0.5$ \\
\midrule
No Defense   &                0.41 &  0.41 &     0.21 &  0.21 \\
\midrule
ALP\cite{DBLP:journals/corr/abs-1803-06373}    &                0.16 &  0.16 &     0.15 &  0.15 \\
IGR\cite{Ross2017Improving}    &                0.37 &  0.37 &     0.33 &  0.33 \\
PAT\cite{Madry2017Towards}   &                0.42 &  0.42 &     0.44 &  0.44 \\
RAT\cite{Araujo2019Robust}    &                0.40 &  0.40 &     0.41 &  0.41 \\

Our \textbf{AT}   &                0.55 &  0.54 &     0.56 &  0.56 \\
\midrule
Our \textbf{AT+ALP} &                \textbf{0.98} &  \textbf{0.98} &     \textbf{0.96} &  \textbf{0.96} \\
\end{tabular}
}

\end{table*}

\subsection{Results and Discussion}

Here, we first present results with \textbf{AT+ALP} on \textbf{17 Flower Category Database}. Compared with previous work, \cite{DBLP:journals/corr/abs-1803-06373} was evaluated under 10-iteration PGD attack and $\epsilon=0.0625$, our work are evaluated under highly challenging PGD attack:the maximum perturbation $\epsilon \in \{0.25,0.5\}$, i.e., $L_\infty \in \{0.25,0.5\}$ with 10 to 200 attack iterations. The bigger the value of $\epsilon$, the bigger the disturbance, the more significant the adversarial image effect is. To the best of our knowledge, such a strong attack has not been previously explored on a wide range of datasets. As shown in Figure  \ref{fig:flowers17:defend-64} that \textbf{our AT+ALP outperform the state-of-the-art in adversarial robustness against highly challenging gray-box and black-box PGD attacks}. For example, under strong 200-iteration \textbf{PGD} gray-box and black-box attacks where prior art has 34\% and 39\% accuracy, our method achieves \textbf{50\%} and \textbf{51\%}.

Table \ref{tab:main_result} shows \textbf{Main Result} of our work: under strong 200-iteration PGD gray-box and black-box attacks, \textbf{our AT+ALP outperform the state-of-the-art in adversarial robustness on all these databases}.

We visualized activation  attention  maps  for  defense  against PGD  attacks.Baseline model is ResNet-101 \cite{DBLP:journals/corr/HeZRS15}, which is pre-trained on \textbf{ImageNet} \cite{ILSVRC15} and fine-tuned on \textbf{17 Flower Category Database} \cite{Nilsback06}, group-0 to group-3 represent the activation attention maps of four groups of convolutional structures in the baseline model, i.e., $conv2\_x$, $conv3\_x$,$conv4\_x$ and $conv5\_x$ of ResNet-101, group-0 extracts of low-level features, group-1 and group-2 extract of mid-level features, group-3 extracts of high-level features\cite{DBLP:journals/corr/ZagoruykoK16a}. We found from Figure \ref{fig:flowers17:attention} that group-0 of \textbf{AT + ALP} can extract the outline and texture of flowers more accurately, and group-3 has a higher level of activation on the whole flower, compared with other defense methods, only \textbf{AT + ALP} makes accurate prediction.

We compared average activations on discriminate parts of \textbf{17 Flower Category Database} for different defense methods.\textbf{17 Flower Category Database} defined  discriminative parts of flowers. So for each image, we got several key  regions which are very important to discriminate its category.Using all testing examples of \textbf{17 Flower Category Database}, we calculated normalized activations on these key regions of these different defense methods. As shown in Table \ref{tab:flower17:at_loss}, \textbf{AT+ALP} got the highest average activations on those key regions, demonstrating that \textbf{AT+ALP} focused on more discriminate features for flowers recognition.

\begin{figure}[htbp]
	\centering
	\subfigure[]{%
			\includegraphics[width=1.5in]{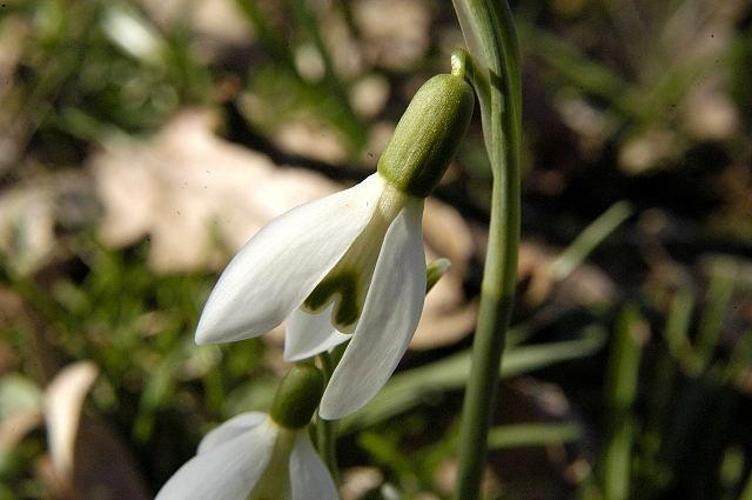}	
			\label{fig:flowers17:demo-a}
	}%
	\subfigure[]{%
			\includegraphics[width=1.5in]{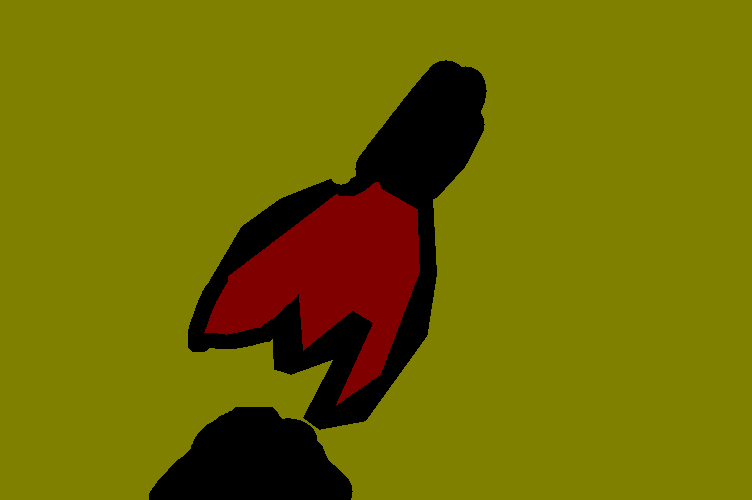}	
			\label{fig:flowers17:demo-b}
	}%

	\centering
	\caption{(a) is original image and (b) is corresponding discriminative parts.\textbf{17 Flower Category Database} defined  discriminative parts of flowers. So for each image, we got several key  regions which are very important to discriminate its category.}
	\label{fig:flowers17:demo}

\end{figure}

\section{Conclusion}
\label{sec:Conclusion}
In this paper, we introduced enhanced defense using a technique we called \textbf{Attention and Adversarial Logit Pairing(AT+ALP)}, a method that encouraged both attention map and logit for pairs of examples to be similar. When being applied to clean examples and their adversarial counterparts, \textbf{AT+ALP} improved accuracy on adversarial examples over adversarial training. Our \textbf{AT+ALP} achieves \textbf{the state of the art} defense on a wide range of datasets against \textbf{PGD} gray-box and black-box attacks.Compared with other defense methods, our \textbf{AT+ALP} is simple and effective, without modifying the model structure, and without adding additional image preprocessing steps.

\begin{figure}[!htb]
	\centering
	\subfigure[No defence]{%
			\includegraphics[width=1.5in]{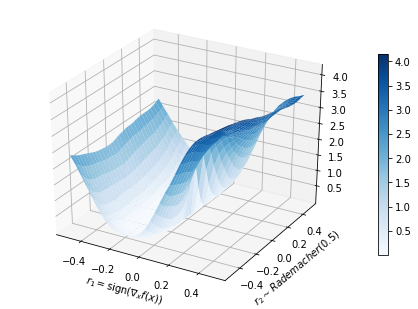}	
			\label{fig:flowers17:loss_landscapes:std-2}
	}%
	\subfigure[ALP]{%
			\includegraphics[width=1.5in]{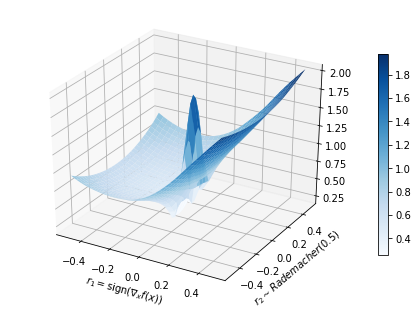}	
			\label{fig:flowers17:loss_landscapes:alp-2}
	}%
	
	\subfigure[AT]{%
			\includegraphics[width=1.5in]{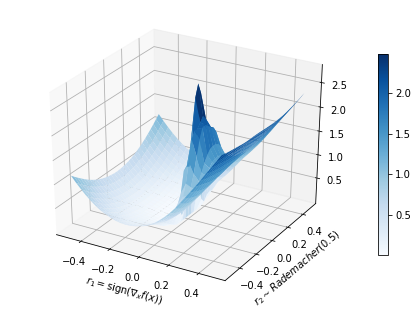}	
			\label{fig:flowers17:loss_landscapes:at-2}
	}%
	\subfigure[AT+ALP]{%
			\includegraphics[width=1.5in]{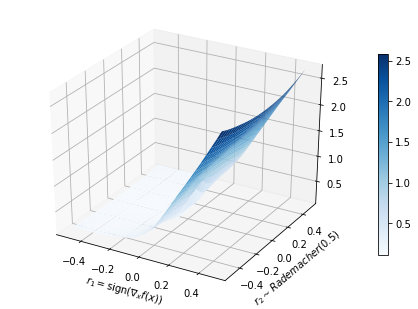}	
			\label{fig:flowers17:loss_landscapes:at-alp-2}
	}%
	
	\centering
	\caption{\textbf{Comparison of loss landscapes}. Loss plots are generated by varying the input to the models, starting from an original input image chosen from the test set. We see that \textbf{ALP} and \textbf{AT} sometimes induces decreased loss near the input locally, and gives a “bumpier” optimization landscape,our \textbf{AT+ALP} has better robustness. The $z$ axis represents the loss. If $x$ is the original input, then we plot the loss varying along the space determined by two vectors: $r1 = sign(\bigtriangledown_xf(x))$ and $r2 \sim Rademacher(0.5)$. We thus plot the following function: $z = loss(x \cdot r1 + y \cdot r2)$.}
	\label{fig:flowers17:loss_landscapes}
\end{figure}


\bibliographystyle{aaai}
\bibliography{mybibliography}

\end{document}